%% file: main.tex
\documentclass[10pt,twocolumn,letterpaper]{article}

\usepackage{iccv}              


\usepackage[noend]{algpseudocode}
\usepackage{algorithm}
\usepackage{graphicx}
\usepackage{comment}
\usepackage{amsmath}

\usepackage{epsfig}
\usepackage{multirow}
\usepackage{boldline}
\usepackage{float}     
\usepackage{adjustbox} 
\usepackage{booktabs}  
\usepackage{caption}
\usepackage{subcaption}
\usepackage{float}
\usepackage{stfloats}
\usepackage[accsupp]{axessibility}  

\definecolor{iccvblue}{rgb}{0.21,0.49,0.74}
\usepackage[pagebackref,breaklinks,colorlinks,allcolors=iccvblue]{hyperref}

\def\paperID{32} 
\def\confName{ICCVW}
\def\confYear{2025}

\title{MK-UNet: Multi-kernel Lightweight CNN for Medical Image Segmentation}

\author{Md Mostafijur Rahman\\
The University of Texas at Austin\\
Austin, Texas\\
{\tt\small mostafijur.rahman@utexas.edu}
\and
Radu Marculescu\\
The University of Texas at Austin\\
Austin, Texas\\
{\tt\small radum@utexas.edu}
}

\begin{document}
\maketitle

\begin{abstract}
In this paper, we introduce MK-UNet, a paradigm shift towards ultra-lightweight, multi-kernel U-shaped CNNs tailored for medical image segmentation. Central to MK-UNet is the multi-kernel depth-wise convolution block (MKDC) we design to adeptly process images through multiple kernels, while capturing complex multi-resolution spatial relationships. MK-UNet also emphasizes the images salient features through sophisticated attention mechanisms, including channel, spatial, and grouped gated attention. Our MK-UNet network, with a modest computational footprint of only 0.316M parameters and 0.314G FLOPs, represents not only a remarkably lightweight, but also significantly improved segmentation solution that provides higher accuracy over state-of-the-art (SOTA) methods across six binary medical imaging benchmarks. Specifically, MK-UNet outperforms TransUNet in DICE score with nearly 333$\times$ and 123$\times$ fewer parameters and FLOPs, respectively. Similarly, when compared against UNeXt, MK-UNet exhibits superior segmentation performance, improving the DICE score up to 6.7\% margins while operating with 4.7$\times$ fewer \#Params. Our MK-UNet also outperforms other recent lightweight networks, such as MedT, CMUNeXt, EGE-UNet, and Rolling-UNet, with much lower computational resources. This leap in performance, coupled with drastic computational gains, positions MK-UNet as an unparalleled solution for real-time, high-fidelity medical diagnostics in resource-limited settings, such as point-of-care devices. Our implementation is available at \url{https://github.com/SLDGroup/MK-UNet}.

\end{abstract}

\input{sections/1.introduction}

\input{sections/2.related_work}
\input{sections/2.technique}
\input{sections/3.experiments}

\input{sections/4.ablations}

\input{sections/5.conclusion}

\section*{Acknowledgments}
This work is supported in part by the NSF grant CNS 2007284, and in part by the iMAGiNE Consortium (https://imagine.utexas.edu/).
\vspace{-0.2cm}

{
    \small
    \bibliographystyle{ieeenat_fullname}
    \bibliography{main}
}

\input{sections/6.supply}

\end{document}

%% file: sections/1.introduction.tex
\section{Introduction}
\label{sec:introduction}

The field of medical image segmentation has experienced transformative growth through the development of U-shaped convolutional neural network (CNN) architectures \cite{ronneberger2015u,oktay2018attention,zhou2018unet++,fan2020pranet} such as UNet \cite{ronneberger2015u}, UNet++ \cite{zhou2018unet++}, AttnUNet \cite{oktay2018attention}, PraNet \cite{fan2020pranet}, UACANet \cite{kim2021uacanet}, and DeepLabv3+ \cite{chen2017deeplab}. These models excel at segmenting medical images, enabling precise segmentation of critical tumors, lesions, or polyps. Attention mechanisms \cite{oktay2018attention,fan2020pranet,woo2018cbam} integrated into these architectures help refine feature maps, thus enhancing pixel-level classification. However, the substantial computational demands of these models, including those with attention mechanisms, limit their applicability in resource-constrained environments such as point-of-care diagnostics.

The introduction of vision transformers \cite{chen2021transunet,cao2021swin,rahman2024g,Rahman_2023_WACV,rahman2023multi,rahman2024emcad,valanarasu2021medical}, including TransUNet \cite{chen2021transunet}, SwinUNet \cite{cao2021swin} and MedT \cite{valanarasu2021medical}, marked a shift towards leveraging self-attention to capture long-range dependencies within images for a comprehensive global view. However, transformers tend to neglect crucial local spatial relationships among pixels which are essential for precise segmentation. Moreover, transformers usually have high memory and computational demands for calculation and fusing attention with convolutional mechanisms, which limits their practical deployment. 

In recent years, a good number of lightweight architectures such as UNeXt \cite{valanarasu2022unext}, CMUNeXt \cite{tang2023cmunext}, MALUNet \cite{ruan2022malunet}, EGE-UNet \cite{ruan2023ege}, and Rolling-UNet \cite{liu2024rolling}, bridge this gap by combining the strengths of CNNs and multi-layer perceptron (MLP). However, most of these architectures are designed for less complex or easy-to-segment applications such as skin lesions, breast cancer in ultrasound, and microscopic cell nuclei/structure segmentation. Consequently, these architectures show poor performance in challenging applications like polyp segmentation due to the high variability in the shape, size, and texture of polyps.

\begin{figure*}[t]
\centering
\begin{subfigure}{.45\textwidth}
  \centering
  \includegraphics[width=1\linewidth]{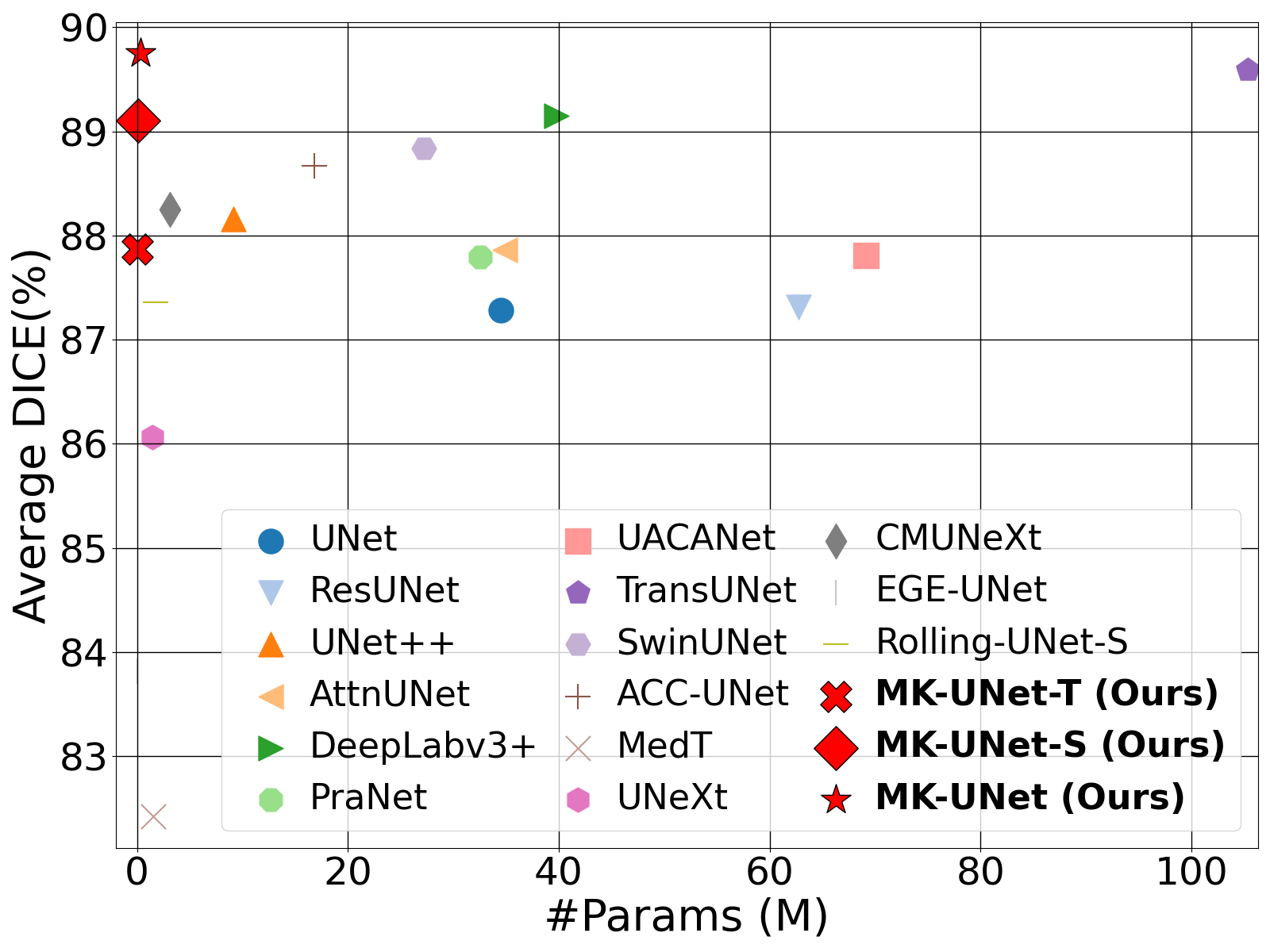}
  \caption{Average DICE scores vs. \#Params}
  \label{fig:dice_vs_params}
\end{subfigure}%
\begin{subfigure}{.45\textwidth}
  \centering
  \includegraphics[width=1\linewidth]{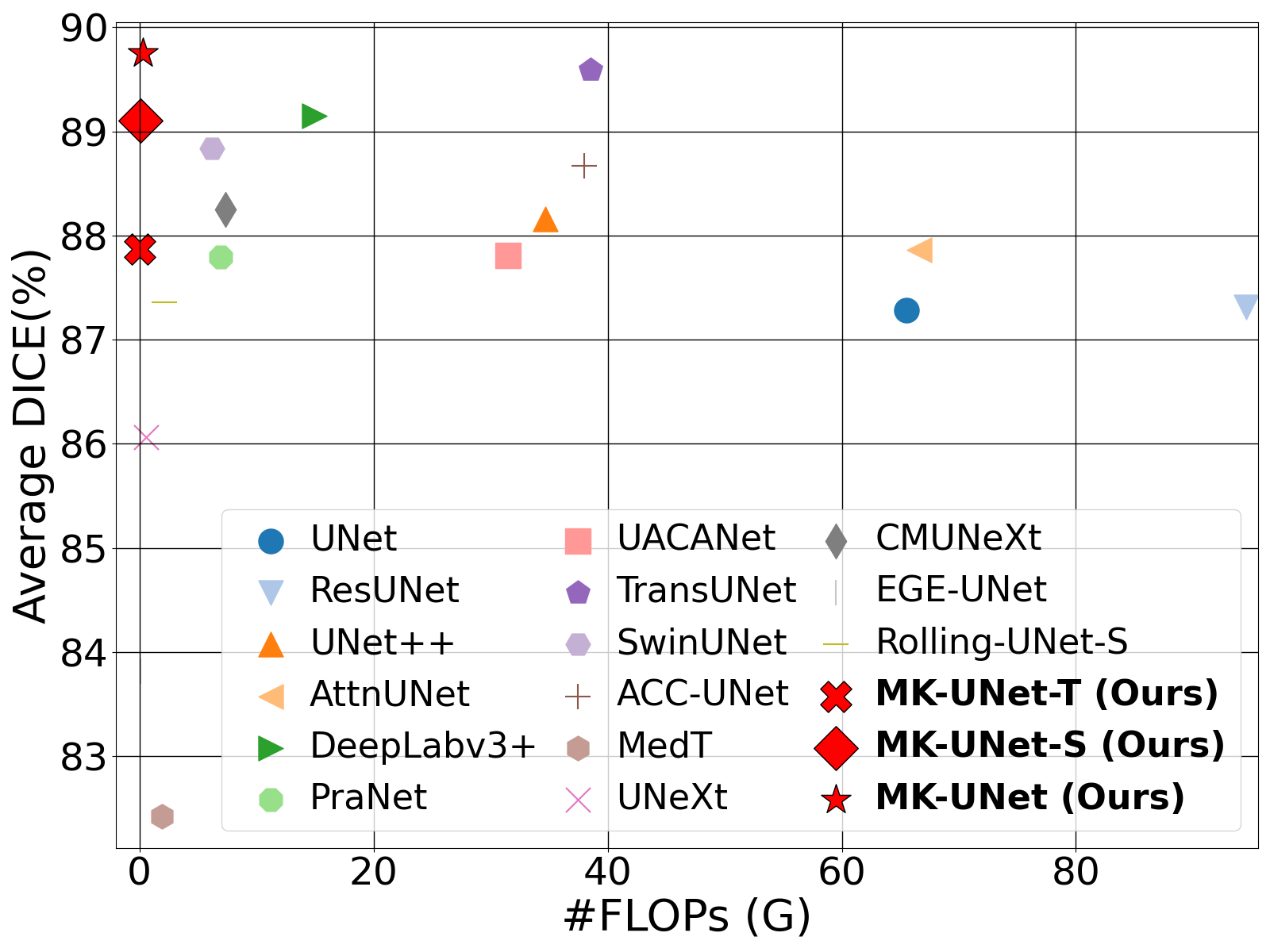}
  \caption{Average DICE scores vs. \#FLOPs}
  \label{fig:dice_vs_flops}
\end{subfigure}
\vspace{-.1cm}
\caption{Comparison of MK-UNet against different SOTA methods over six binary medical image segmentation datasets. As shown, our MK-UNet has the second lowest \#Params and \#FLOPs (behind EGE-UNet \cite{ruan2023ege}), yet the highest DICE scores. Though EGE-UNet has the lowest \#Params and \#FLOPs among existing methods, it has the lowest DICE score as well.}
\label{fig:dice_vs_params_flops}
\vspace{-.2cm}
\end{figure*}


To address these computational and precision challenges, we introduce MK-UNet, a significant breakthrough in image segmentation, which leverages the benefits of a multi-kernel perspective (i.e., $k_1=k_2$ or $k_1 \neq k_2$ for $k_1, k_2 \in Kernels$) and depth-wise convolutions to address the computational complexity and challenges inherent in existing CNN- and transformer-based models. Depth-wise convolutions drastically reduce the computational load, making the network more efficient without sacrificing the ability to capture detailed features within the image. Additionally, our multi-kernel property enables the model to effectively handle feature representations at \textit{same or varying} receptive fields, thus allowing for a more robust and comprehensive analysis of complex images in diverse applications. Moreover, by incorporating sophisticated convolutional multi-focal (channel and spatial) attention mechanisms \textit{only} in our decoder further refines the feature maps by capturing the image salient features. We note that our network is effective for segmentation in both scenarios, whether the regions of interest vary significantly in size and shape or remain relatively uniform. By integrating these new ideas, MK-UNet achieves a fine balance between computational efficiency and segmentation accuracy, thus offering an ultra-lightweight model that not only surpasses the performance of heavyweight counterparts (in DICE scores), but it does so with significantly fewer \#Params and \#FLOPs. Our contributions are as follows:

\begin{itemize}
    \item \textbf{New Lightweight Multi-kernel UNet:} We propose a new end-to-end U-shaped network, MK-UNet, for medical image segmentation, which encodes an image using lightweight multi-kernel convolutions to capture multi-resolution spatial representations. MK-UNet also progressively refines the multi-resolution spatial representations using multi-kernel convolutional attention. Of note, our MK-UNet-T (tiny) has only 0.027M and 0.062G \#Params and \#FLOPs, respectively, yet provides SOTA performance. Moreover, MK-UNet (standard) has only 0.316M \#Params and 0.314G \#FLOPs. The extremely low model size (\#Params) and computations (\#FLOPs) make our MK-UNet easy to deploy in point-of-care diagnostics or resource-constraint environments (e.g., mobile or edge devices).
    \item \textbf{Lightweight Multi-kernel Inverted Residual:} We introduce MKIR, a new Multi-Kernel Inverted Residual block that performs depth-wise convolutions with multiple kernels (i.e., $k_1=k_2$ or $k_1 \neq k_2$, for $k_1, k_2 \in Kernels$). Our encoder extracts features using the MKIR block; this choice is motivated by the need to efficiently process and encode diverse and complex structures in medical images, thus providing a rich representation with minimal computational costs.
    \item \textbf{Lightweight Multi-kernel Inverted Residual Attention:} We propose Multi-Kernel Inverted Residual Attention (MKIRA), a new block to refine and enhance multi-scale salient features by suppressing irrelevant regions. In our decoder, MKIRA enhances features discrimination by focusing on key feature channels and highlighting the important spatial regions in an image. This ensures that the decoder can reconstruct precise and accurate segmentation maps by focusing only on the most critical aspects of the encoded features. 
    \item \textbf{Improved Performance across Various Tasks and Benchmarks:} We empirically show that MK-UNet significantly improves the performance of medical image segmentation compared to SOTA methods with a significantly lower computational cost (as shown in Fig. \ref{fig:dice_vs_params_flops}) on six binary medical image segmentation benchmarks that belong to four different tasks.
\end{itemize}

The remaining of this paper is organized as follows: Section \ref{sec:method} describes the proposed method. Section \ref{sec:experiments} explains our experimental setup and results on six medical image segmentation benchmarks. Section \ref{sec:ablation_study} covers different ablation experiments. Lastly, Section \ref{sec:conclusion} concludes the paper. 

%% file: sections/2.related_work.tex
\section{Related Work}
\label{sec:related_work}


\subsection{Convolutional Neural Networks (CNNs)}
The advent of CNNs marks a significant shift in medical image segmentation \cite{ronneberger2015u, oktay2018attention, zhou2018unet++, fan2020pranet, kim2021uacanet, chen2017deeplab}. Pioneering work such as Fully Convolutional Networks (FCNs) \cite{long2015fully} laid the foundation for end-to-end segmentation models. FCNs replace fully connected layers with convolutional layers, thus enabling pixel-wise predictions and efficient learning of spatial hierarchies in images. U-Net \cite{ronneberger2015u} became a cornerstone in medical image segmentation due to its encoder-decoder architecture with skip connections. This design effectively combines high-resolution features from the encoder with context information from the decoder, hence leading to precise segmentation even with limited training data. The sophisticated design of U-shaped architecture for pixel-level segmentation tasks motivates us to choose the U-shaped design in our proposed network.

U-Net's success has inspired numerous variants and improvements. Inspired by residual learning in ResNet \cite{he2016deep}, ResUNet \cite{zhang2018road} uses residual blocks to facilitate gradient flow and improve convergence, addressing the vanishing gradient problem in deep networks. Zhou et al.\cite{zhou2018unet++} introduce UNet++, which uses dense nested skip connections to further enhance the feature propagation and improve the segmentation accuracy. AttnUNet \cite{oktay2018attention} incorporates attention mechanisms to focus on relevant regions in the feature maps, improving segmentation performance by suppressing irrelevant background noise. Fan et al. \cite{fan2020pranet} introduce PraNet for precise polyp segmentation that employs parallel reverse attention and edge-guidance to refine segmentation boundaries. UACANet \cite{kim2021uacanet} uses uncertainty-aware mechanisms to improve the reliability and robustness of segmentation outcomes. DeepLabv3+ \cite{chen2017deeplab} integrates atrous convolutions and spatial pyramid pooling to capture multi-scale context information. ACC-UNet \cite{ibtehaz2023acc} employs adaptive context capture mechanisms to dynamically adjust the receptive fields based on the input image.

\subsection{Vision Transformers}
Vision Transformers (ViTs) \cite{dosovitskiy2020image, liu2021swin} have emerged as a powerful alternative to CNNs, offering a new paradigm for medical image analysis tasks by leveraging the self-attention mechanism \cite{chen2021transunet, cao2021swin, rahman2024emcad, Rahman_2023_WACV, rahman2023multi, rahman2024g, valanarasu2021medical}. By combining the strengths of CNNs for local feature extraction and Transformers for capturing long-range dependencies, TransUNet \cite{chen2021transunet} achieves superior performance in medical image segmentation. SwinUNet \cite{cao2021swin} is introduced based on the Swin Transformer \cite{liu2021swin} architecture, which utilizes shifted windows to achieve hierarchical feature representation, enabling efficient computation. MedT \cite{valanarasu2021medical}, a lightweight Transformer model specifically designed for medical image segmentation, which employs gated axial attention mechanisms to focus on relevant regions and reduce computational complexity. Rahman et al. introduce CASCADE \cite{Rahman_2023_WACV}, a cascaded-attention decoding network using standard convolutions. Recently, EMCAD \cite{rahman2024emcad} introduces a depthwise convolutions-based multi-scale decoder. Although, CASCADE and EMCAD perform well in medical image segmentation, their segmentation accuracy and computational complexity solely depend on the strength and complexity of the exiting pretrained transformer encoder they use, thus making them less suitable for resource-constrained settings. In contrast, we propose to design an extremely efficient (ultra-lightweight) end-to-end (both encoder and decoder) architecture using the multi-kernel trick (where, $k_1=k_2$ or $k_1 \neq k_2$ for $k_1, k_2 \in Kernels$) with depthwise convolutions.

\subsection{Lightweight CNNs}
Recent efforts have focused on making CNNs more efficient for real-time and resource-constrained environments. MobileNets \cite{howard2017mobilenets} and EfficientNets \cite{tan2019efficientnet} introduce depthwise separable convolutions and compound scaling, respectively, to create lightweight models with competitive performance. Additionally, several novel lightweight architectures have been developed to further enhance the efficiency of medical image segmentation \cite{valanarasu2022unext, tang2023cmunext, ruan2023ege, liu2024rolling}. UNeXt \cite{valanarasu2022unext} uses hybrid convolutional and transformer blocks to capture local and global features efficiently, improving segmentation accuracy while maintaining computational efficiency. CMUNeXt \cite{tang2023cmunext} combines convolutional and multi-scale features to enhance segmentation performance. EGE-UNet \cite{ruan2023ege} integrates edge-guided mechanisms to refine segmentation boundaries. Rolling-UNet \cite{liu2024rolling} incorporates rolling convolutional blocks to enhance the model's ability to capture long-range dependencies.

%% file: sections/2.technique.tex
\begin{figure*}[t]
\centering
\includegraphics[width=0.88\textwidth]{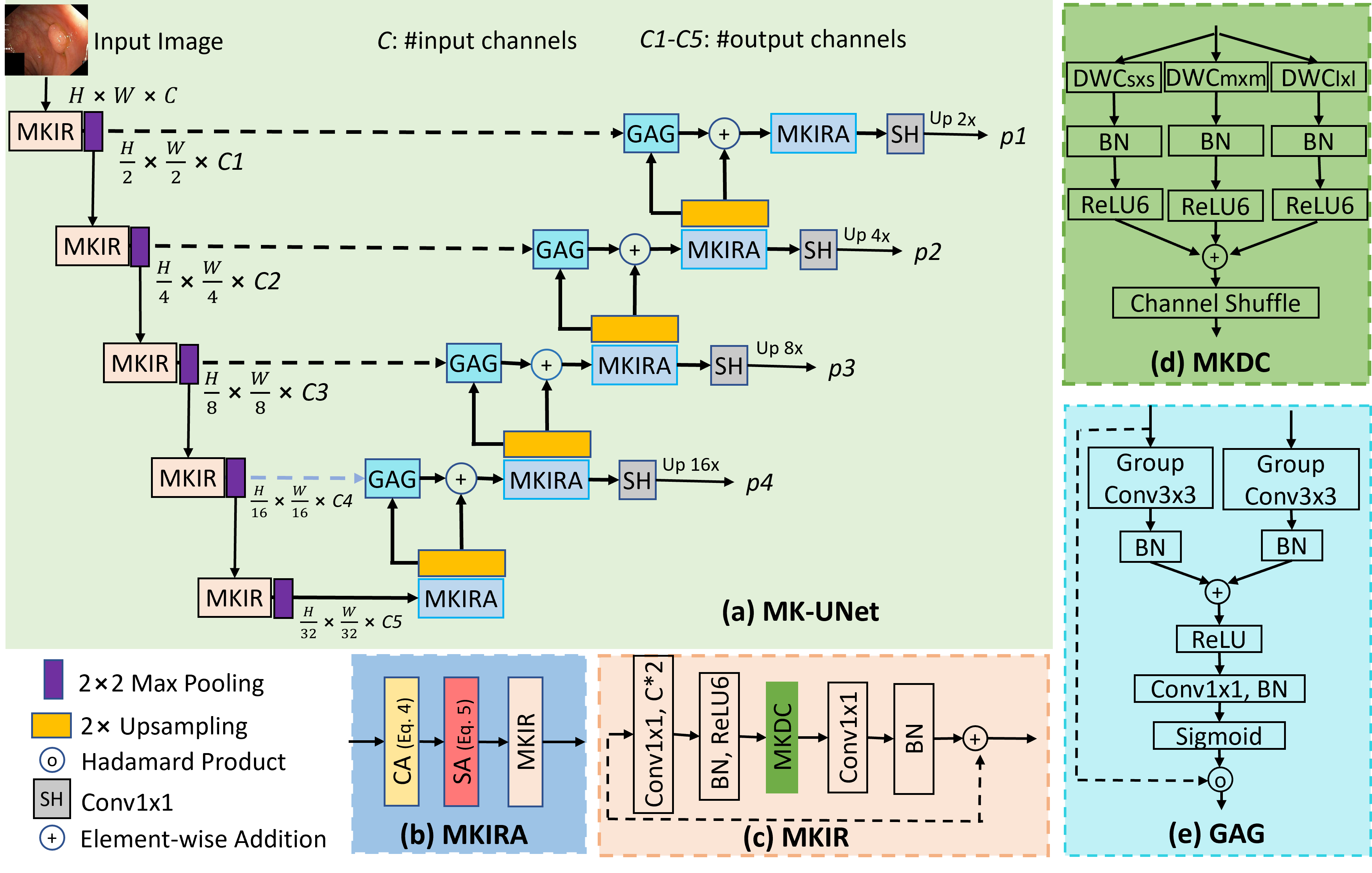}
\vspace{-.2cm}
\caption{The proposed network diagram. (a) Five-stage MK-UNet network (b) Multi-kernel inverted residual attention (MKIRA), (c) Multi-kernel inverted residual (MKIR), (d) Multi-kernel (parallel) depth-wise convolution (MKDC), (e) Grouped attention gate (GAG). p1, p2, p3, and p4 are output segmentation maps. \textbf{Note}: Channel attention (CA), Spatial attention (SA).}
\label{fig:architecture}
\vspace{-.2cm}
\end{figure*}

\section{Method}
\label{sec:method}
We describe next our core building blocks. Then, we introduce our MK-UNet architecture by integrating these blocks into the baseline UNeXt \cite{valanarasu2022unext} (Fig. \ref{fig:architecture}a in green box). 

\subsection{Multi-kernel Inverted Residual (MKIR)} 
We first introduce the multi-kernel inverted residual ($MKIR$) block to generate and refine feature maps (Fig. \ref{fig:architecture}c). By utilizing multiple (same or different) kernel sizes, $MKIR$ allows for better understanding of both fine-grained details and broader contexts, thereby enabling a comprehensive representation of the input. As shown in Fig. \ref{fig:architecture}c, the process begins by expanding the \#channels (i.e., expansion\_factor = 2) through point-wise convolution $PWC_1$, batch normalization $BN$ \cite{ioffe2015batch}, and $ReLU6$ activation \cite{krizhevsky2010convolutional}. This is followed by multi-kernel depth-wise convolution $MKDC$ for capturing application-specific complex spatial contexts. A subsequent point-wise convolution $PWC_2$ and $BN$ restore the original \#channels. The MKIR (Eq. \ref{eq:mkir}) significantly reduces the computational cost while ensuring rich feature representation: 
\begin{equation}
\scalebox{0.75}{\(
    MKIR(x) = BN(PWC_2(MKDC(ReLU6(BN(PWC_1(x))))))
\)}
    \label{eq:mkir}
\end{equation}
where $MKDC$ for multiple kernels ($K$) is defined in Eq. \ref{eq:mkdc} and Fig. \ref{fig:architecture}d: 
\begin{equation}
    \scalebox{0.9}{\( 
    MKDC(x) = CS(\sum_{k \in K} DWCB_{k}(x))
    \)}
    \label{eq:mkdc}
\end{equation}
where $DWCB_{k}(x) = ReLU6(BN(DWC_{k}(x)))$. Here, $DWC_{k}$(.) is a depth-wise convolution with the kernel $k\times k$. To address the channel independence in depth-wise convolution, a channel shuffle ($CS$) is used to ensure the inter-channel information flow. Our $MKDC(.)$ differs from $MSCB$ (in EMCAD \cite{rahman2024emcad}) in their core theoretical concepts. Our multi-kernel trick supports both $k_1=k_2$ (same-size kernels) and $k_1 \neq k_2$ (different-size kernels) for $k_1, k_2 \in K$ versus conventional multi-scale (only $k_1 \neq k_2$) designs \cite{rahman2024emcad, lin2023scale, seo2022spatial}, thus allowing adaptable context extraction.
This conceptual distinction allows MK-UNet to adapt kernel sizes based on application-specific needs (e.g., large kernels for large objects, small kernels for small objects, or mixed for both objects segmentation).  

\subsection{Multi-kernel Inverted Residual Attention (MKIRA)}
We present a lightweight multi-kernel inverted residual attention, MKIRA, to refine the feature maps ($x$). MKIRA uses channel attention ($CA$) \cite{hu2018squeeze} to focus on relevant channels, a spatial attention ($SA$) \cite{chen2017sca} to capture the local context, and a multi-kernel inverted residual ($MKIR$) to enrich the feature maps while capturing contextual relationships. The $MKIRA$ (Fig. \ref{fig:architecture}b) is given in Eq. \ref{eq:mkira}: 
\begin{equation}
    MKIRA(x) = MKIR(SA(CA(x)))
    \label{eq:mkira}
\end{equation}


\noindent \textit{Channel Attention (CA):} 
We use $CA$ \cite{hu2018squeeze} to enhance relevant features by applying adaptive max ($AMP$) and average pooling ($AAP$) to condense the spatial information \cite{Rahman_2023_WACV}; this is followed by channel reduction ($r = 16$) through point-wise convolution $PWC_1$ with $ReLU$ ($R$) activation \cite{nair2010rectified} and expansion through $PWC_2$. The $Sigmoid$ ($\sigma$) activation generates attention weights, which are then applied to the input via the Hadamard product ($\circledast$), while focusing on crucial feature maps. The $CA$ is defined in Eq. \ref{eq:ca}:
 \begin{equation}
 \scalebox{0.62}{\( CA(x)=\sigma(PWC_2(R(PWC_1(AMP(x))))+PWC_2(R(PWC_1(AAP(x))))) \circledast x
 \)}
    \label{eq:ca}
\end{equation}

\noindent \textit{Spatial Attention (SA):} 
We use $SA$ \cite{chen2017sca} to focus on specific image regions to highlight key features, crucial for accurate segmentation. The $SA$ aggregates maximum ($Ch_{max}$) and average ($Ch_{avg}$) channel values to highlight local details, then it employs a large-kernel ($7 \times 7$) convolution ($LKC$) to strengthen the contextual connections. The $Sigmoid$ ($\sigma$) activation derives attention weights, applied to the input $x$ via the Hadamard product ($\circledast$), ensuring targeted refinement. The $SA$ is derived in Eq. \ref{eq:sa}:
\begin{equation}
    SA(x) = \sigma(LKC([Ch_{max}(x), Ch_{avg}(x)])) \circledast x
    \label{eq:sa}
\end{equation}
\normalsize

\subsection{Grouped attention gate (GAG)}
We use a grouped attention gate ($GAG$, Fig. \ref{fig:architecture}e) that mixes the feature maps with the attention coefficients for enhancing the relevant features and suppressing the irrelevant ones. By utilizing a gating signal from higher-resolution features, $GAG$ directs the information flow, thus improving medical image segmentation accuracy. Unlike Attention UNet \cite{oktay2018attention}, which processes signals with $1\times1$ convolution, our method applies $3\times3$ group convolutions ($GC$) to both gating ($g$) and input ($x$) feature maps separately \cite{rahman2024emcad}. After convolution, the features undergo batch normalization ($BN$) and get combined via addition, followed by $ReLU$ ($R$) activation. Subsequently, a $1\times1$ convolution and batch normalization ($BN$) produce a unified feature map which, after the $Sigmoid$ activation ($\sigma$), generates the attention coefficients. These coefficients adjust the input feature $x$, and create an attention-enhanced output. $GAG$ is defined in Eq.s \ref{eq:ags_1}:
\begin{equation}
\scalebox{0.7}{\(
   GAG(g, x) = x \circledast \sigma(BN(Conv(R(BN(GC_g(g) + BN(GC_x(x)))))))))
   \)}
    \label{eq:ags_1}
\end{equation} 




\subsection{Multi-kernel UNet (MK-UNet)}
Our MK-UNet employs a multi-kernel approach across five encoding and decoding stages to generate high-resolution segmentation maps, as depicted in Fig. \ref{fig:architecture}a. Each encoding stage uses a multi-kernel inverted residual (MKIR) block to produce $C_i$ feature maps, followed by max pooling for downsampling while retaining crucial information. The output from the final encoding stage passes through a multi-kernel inverted residual attention (MKIRA) block in the decoder's initial stage, significantly refining the feature maps. These are then upsampled using bilinear interpolation for subsequent decoding stages. Decoder stages integrate skip-connections with refined features using a grouped attention gate (GAG) followed by additive aggregation. The resultant feature maps are refined through the MKIRA block and up-sampled (only bilinear $2\times$, no convolutions) to align with the later stages. 

The segmentation heads (SHs) at the last four stages output the segmentation maps $p_4$, $p_3$, $p_2$, and $p_1$. We consider the feature map $p_1$ as final prediction and obtain the final segmentation output by employing a $Sigmoid$ for binary segmentation or a $Softmax$ activation for multi-class segmentation. We optimize the loss of only final prediction $p_1$ for all binary segmentation tasks. However, we recommend using deep supervision for multi-class segmentation.

%% file: sections/3.experiments.tex
\section{Experiments and Results}
\label{sec:experiments}
\subsection{Datasets}
\label{assec:Datasets}
We evaluate the MK-UNet's efficacy across six datasets covering four segmentation tasks, including breast cancer (BUSI \cite{al2020dataset}, 647 images: 437 benign and 210 malignant), polyp (ClinicDB \cite{bernal2015wm} with 612 images, and ColonDB \cite{vazquez2017benchmark} with 379 images), skin lesion (ISIC18 \cite{codella2019skin}, 2,594 images), and cell nuclei/structure segmentation (DSB18 \cite{caicedo2019nucleus} with 670 images, and EM \cite{cardona2010integrated} with 30 images). These datasets, collected from various imaging centers, offer a broad diversity in image characteristics, ensuring a comprehensive evaluation. An 80:10:10 train-val-test split was applied across all datasets and the DICE score of testset is reported.


\subsection{Implementation details}
\label{ssec:impl_details}

Our networks are developed and evaluated using Pytorch 1.11.0, operating on a single NVIDIA RTX A6000 GPU equipped with 48GB of RAM.  We utilize multi-scale kernels $[1,3,5]$ within our MKDC, based on an ablation study. The architecture employs a series of parallel depth-wise convolutions in the MK-UNet network, standardizing on channel configurations of $[16,32,64,96,160]$ across all experiments, unless specified otherwise. Model optimization is achieved via the AdamW \cite{loshchilov2017decoupled} optimizer with both learning rate and weight decay set to $1e-4$. Training spans over 200 epochs with batches of 16, during which we save the model achieving the highest DICE score. 

Image dimensions are set to $256\times256$ pixels for BUSI \cite{al2020dataset}, ISIC18 \cite{codella2018skin}, EM \cite{cardona2010integrated}, and DSB18 \cite{caicedo2019nucleus} datasets, while for ClinicDB \cite{bernal2015wm} and ColonDB \cite{vazquez2017benchmark}, the resolution is adjusted to $352\times352$ pixels. We utilize a multi-scale training approach, with scales of \{0.75, 1.0, 1.25\}, and enforce gradient clipping at 0.5. We do not apply any form of augmentation and use a hybrid loss function that combines (1:1) weighted BinaryCrossEntropy (BCE) with a weighted Intersection over Union (IoU) loss.




\begin{table*}[t]
\begin{center}
\begin{adjustbox}{width=0.88\textwidth}
\begin{tabular}{l|c|r|r|r|r|r|rr|rr|r}
\toprule
\multirow{2}{*}{Network} & \multirow{2}{*}{Pretrain} & \multirow{1}{*}{\#Params} & \multirow{1}{*}{FLOPs} & \multirow{1}{*}{Throughput} & \multirow{2}{*}{BUSI} &  \multirow{2}{*}{ISIC18} & \multicolumn{2}{c|}{Polyp} & \multicolumn{2}{c|}{Cell} & \multirow{2}{*}{Avg.}\\
\cline{8-11}
& & (M) & (G) & (/s) & & & Clinic & Colon & DSB18 & EM & \\
\midrule
UNet \cite{ronneberger2015u} & No & 34.53M & 65.53G & 119.03 & 74.04 & 86.67 & 91.43 & 83.95 & 92.23 & 95.36 & 87.28 \\
UNet++ \cite{zhou2018unet++} & No & 9.16M & 34.65G & 136.98 & 74.76 & 87.46 & 91.52 & 87.88 & 91.97 & 95.38 & 88.16 \\
AttnUNet \cite{oktay2018attention} & No & 34.88M & 66.64G & 108.68 & 74.48 & 87.05 & 91.50 & 86.46 & 92.22 & 95.45 & 87.86 \\
DeepLabv3+ \cite{chen2017deeplab} & Yes & 39.76M & 14.92G & 128.20 & 76.81 & 88.64 & 92.46 & 89.86 & 92.14 & 94.96 & 89.15 \\
PraNet \cite{fan2020pranet} & Yes & 32.55M & 6.93G & 64.10 & 75.14 & 88.46 & 91.71 & 89.16 & 89.89 & 92.37 & 87.79 \\
UACANet \cite{kim2021uacanet} & Yes & 69.16M & 31.51G & 43.28 & 76.96 & 88.72 & 93.29 & 89.76 & 88.86 & 89.28 & 87.81 \\
TransUNet \cite{chen2021transunet} & Yes & 105.32M & 38.52G & 65.35 & 78.01 & 89.04 & 93.18  & 89.97 & 92.04 & 95.27 & 89.59 \\
SwinUNet \cite{cao2021swin} & Yes & 27.17M  & 6.2G & 80.64 & 77.38 & 88.66 & 92.42 & 89.07 & 91.03 & 94.47 & 88.84 \\
Rolling-UNet-S \cite{liu2024rolling} & No & 1.78M &  2.1G & 57.14 & 76.38 &  87.35 & 90.23 & 82.48 & 92.50 & 95.23 & 87.36\\
MedT \cite{valanarasu2021medical} & No & 1.57M & 1.95G & 8.40 & 69.23 & 86.78 & 83.44 & 68.90 & 92.28 & 93.87 & 82.42 \\
UNeXt \cite{valanarasu2022unext} & No & 1.47M & 0.57G & 172.41 & 74.71 & 87.78 & 90.20 & 83.84 & 86.01 & 93.81 & 86.06 \\
CMUNeXt \cite{tang2023cmunext} & No & 0.418M & 1.09G & \textbf{175.43} & 77.34 & 87.51 & 92.82 & 83.85 & 92.58 & 95.38 & 88.25 \\
EGE-UNet \cite{ruan2023ege} & No &  0.054M &  0.072G & 101.01 & 71.34 &  86.95 & 84.76 & 76.03 & 90.10 & 93.76 & 83.82 \\
UltraLight\_VM\_UNet \cite{wu2024ultralight} & No & 0.050M & \textbf{0.060G} & 98.03 & 72.31 & 87.85  & 87.11  & 80.06 & 91.88 & 93.96 & 85.53 \\
\midrule
\textbf{MK-UNet-T (Ours)} & No & \textbf{0.027M} & 0.062G & 140.85 & 75.64 & 88.19 & 91.26 & 85.03 & 92.38 & 94.69 &  87.87 \\
\textbf{MK-UNet-S (Ours)} & No & 0.093M & 0.125G & 139.42 & 77.26 & 88.57 & 92.31 & 88.78 & 92.45 & 95.22 &  89.10 \\
\textbf{MK-UNet (Ours)} & No & 0.316M & 0.314G & 138.89 & 78.04 & 88.74 & 93.48 & 90.01 & 92.71 & 95.52 &  89.75 \\
\textbf{MK-UNet-M (Ours)}    & No  & 1.15M  & 0.951G &  133.73 & 78.27 & 89.08 & 93.67 & 90.27 & 92.74 & 95.62 & 89.94 \\
\textbf{MK-UNet-L (Ours)}    & No    & 3.76M & 3.19G &  122.62 & \textbf{79.02} & \textbf{89.25} & \textbf{93.85} & \textbf{91.82} & \textbf{92.80} & \textbf{95.67} & \textbf{90.40} \\

\bottomrule
\end{tabular}
\end{adjustbox}
\vspace{-0.2cm}
\caption{Results of binary (breast cancer, skin lesion, polyp, and cell) segmentation. We reproduce the results of SOTA methods using their publicly available implementations with our 80:10:10 train-val-test splits. FLOPs of all methods are reported for $256\times256$ inputs. The FLOPs of all methods for polyp segmentation with $352\times352$ inputs will be higher. We report DICE scores (\%) averaging over five runs, thus having 1-4\% standard deviations. $[C1, C2, C3, C4, C5]$ = MK-UNet-T $[4, 8, 16, 24, 32]$, MK-UNet-S $[8, 16, 32, 48, 80]$, MK-UNet $[16, 32, 64, 96, 160]$, MK-UNet-M $[32, 64, 128, 192, 320]$, and MK-UNet-L $[64, 128, 256, 384, 512]$. Best results are shown in \textbf{bold}.}
\label{tab:results_binary_mis}
\end{center}
\vspace{-0.5cm}
\end{table*}

\subsection{Results}
\label{ssec:results}
Table \ref{tab:results_binary_mis} and Fig. \ref{fig:dice_vs_params_flops} compare our MK-UNet with SOTA CNNs and transformers on six datasets of four binary medical segmentation tasks. Our MK-UNet achieves the top average DICE score of 89.75\%, while maintaining an ultra-lightweight footprint of only 0.316M \#Params and 0.314G \#FLOPs. Our MK-UNet-T with 0.027M \#Params and 0.062G \#FLOPs, outperforms the existing most tiny model EGE-UNet \cite{ruan2023ege} by on an average 5.93\% DICE score over six datasets. The multi-kernel depth-wise convolutions within our network, alongside local attention mechanisms, play a crucial role in these strong results. Our MK-UNet's performance on different datasets highlights its superior ability to balance accuracy with computational efficiency, setting a new benchmark for point-of-care diagnostics. The quantitative results of four different tasks are described next.

\subsubsection{Breast cancer segmentation}
We focus on breast cancer segmentation in ultrasound images using the BUSI dataset. Our MK-UNet model notably outperforms existing methods, achieving the highest DICE score of 78.04\% with remarkably low computational requirements. This achievement underscores the efficiency and effectiveness of our approach, particularly when compared against more complex models. For instance, TransUNet, despite having substantially more parameters (105.32M) and FLOPs (38.52G), achieves a lower DICE score of 78.01\%. Similarly, ACC-UNet, with its 16.8M parameters and 38.0G FLOPs, closely follows with a DICE score of 77.02\%. Additionally, compared to the computationally comparable model UNeXt, our MK-UNet demonstrates a significant improvement, surpassing it by a margin of 3.33\% with 4.7$\times$ fewer \#Params. 

\subsubsection{Skin lesion segmentation}
On the ISIC18 dataset for skin lesion segmentation, our MK-UNet network exhibits a commendable performance with a DICE score of 88.74\%. This result not only showcases our model's robustness, but also its capability to handle the complex variations present in skin lesion images. We note that TransUNet leads with a marginally higher score of 89.04\% but with a significantly higher cost (333$\times$ and 123$\times$ larger \#Params and \#FLOPs, respectively). Our MK-UNet achieves near-top performance while maintaining minimal computational requirements—merely 0.316M \#Params and 0.314G \#FLOPs. 

\subsubsection{Polyp segmentation}
\label{sssec:polyp_results}
In polyp segmentation on Clinic and Colon datasets, our MK-UNet model excels with leading scores of 93.48\% and 90.01\%, respectively, and with a remarkably low computational footprint. This performance surpasses notable competitors like DeepLabv3+ and ACC-UNet, which, despite their higher resource consumption, do not match MK-UNet's precision. Specifically, DeepLabv3+ reaches 92.46\% on Clinic and 89.86\% on Colon with 126$\times$ and 47.5$\times$ larger parameters and FLOPs, respectively, while ACC-UNet closely follows but still falls short. 

\subsubsection{Microscopic cell nuclei/structure segmentation}
For cell nuclei/structure segmentation on the DSB18 and EM datasets, our MK-UNet network demonstrates exceptional accuracy, achieving DICE scores of 92.71\% and 95.52\%, respectively. In contrast, other models like TransUNet and UNeXt, despite their heavy-weight design and higher computational demands, do not surpass MK-UNet's DICE score. For instance, TransUNet achieves lower scores of 92.04\% on DSB18 and 95.27\% on EM, while UNeXt falls 6.70\% behind MK-UNet.

\begin{figure*}[t]
\centering
\includegraphics[width=0.95\textwidth]{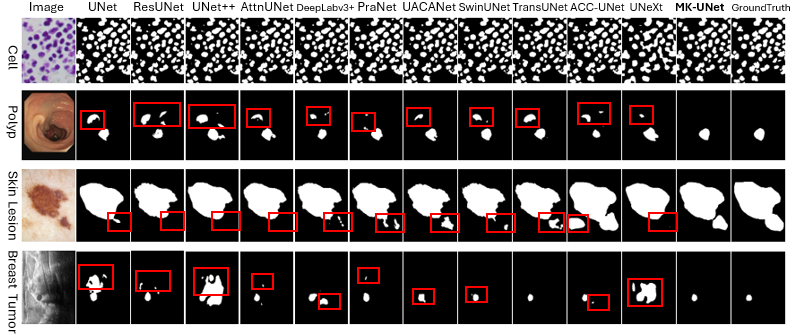}
\vspace{-.2cm}
\caption{Qualitative results of our MK-UNet and SOTA methods. The incorrect segmented regions by different methods are highlighted using the red rectangular box.} 
\label{fig:qualitative}
\vspace{-.2cm}
\end{figure*}

\begin{table*}[t]
\begin{center}
    \begin{adjustbox}{width=0.75\textwidth}
\begin{tabular}{c|r|r|r|r|r|r|r|r}
\toprule
Network     & \#Params & \#FLOPs & BUSI & Clinic & Colon & ISIC18 & DSB18 & EM                    \\
\midrule
Baseline UNeXt          & 1.47M & 0.57G & 74.71 &  90.20 & 83.84 &  87.78 & 86.01 & 93.81    \\
Redesigned Mobile UNet          & 0.271M     & 0.230G     & 72.41  & 90.90 & 84.15 & 87.20 & 90.52 & 94.87  \\
MKIR          & 0.306M    & 0.300G     & 74.74 & 92.63 & 86.46 & 88.22 & 92.40 & 95.31         \\
MKIR + GAG          & 0.310M     & 0.311G    & 74.98  & 91.97 & 86.56 & 88.34 & 92.67 & 95.48     \\
MKIR + MKIRA         & 0.311M    & 0.303G    &  76.61 & 92.64 & 89.40 &  88.56 & 92.64 & 95.37     \\
\textbf{MKIR + GAG + MKIRA (Ours)}        & 0.316M    & 0.314G    & \textbf{78.04} & \textbf{93.48} & \textbf{90.01} & \textbf{88.64} & \textbf{92.71} & \textbf{95.52}       \\
\bottomrule 
\end{tabular}
\end{adjustbox}
\end{center}
\vspace{-0.5cm}
\caption{Effect of different components of MK-UNet with \#channels = $[16,32,64,96,160]$ and $[1,3,5]$ kernels. UNeXt has \#channels = $[16,32,128,160,256]$. We design Mobile UNet following the structure of UNeXt network. However, we use the \#channels = $[16,32,64,96,160]$ and kernel size of $[3]$ with the original inverted residual block (IRB) in the Mobile UNet. We report the DICE scores (\%) averaging over five runs. Best results are shown in \textbf{bold}.}
\vspace{-0.4cm}
\label{tab:ablation_components}
\end{table*}

\begin{table*}[t]
\centering 
    {
\begin{adjustbox}{width=0.8\textwidth}
{\begin{tabular}{crrr|crrr}
\toprule
Convolution kernels & \#Params(M) & FLOPs(G) & DICE & Convolution kernels & \#Params(M) & FLOPs(G) & DICE\\
\midrule
$1\times1$           & 0.272 & 0.220 & 70.83 & $5\times5$           & 0.299 & 0.276 & 76.81 \\
$1\times1, 1\times1$         & 0.275 & 0.229 & 71.11 & $1\times1, 5\times5$         & 0.303 & 0.286 & 77.05 \\
$3\times3$           & 0.281 & 0.239 & 76.42 & $3\times3, 3\times3, 3\times3$       & 0.306 & 0.295 & 76.86 \\
$1\times1, 3\times3$         & 0.284 & 0.248 & 76.81 & $3\times3, 5\times5$         & 0.312 & 0.304 & 77.62 \\
$1\times1, 1\times1, 3\times3$       & 0.288 & 0.257 & 77.08 & \textbf{1$\times$1, 3$\times$3, 5$\times$5}       & 0.316 & 0.314 & \textbf{78.04} \\
$3\times3, 3\times3$         & 0.294 & 0.267 & 76.83 & $5\times5, 5\times5$         & 0.331 & 0.342 & 77.88 \\
$1\times1, 3\times3, 3\times3$       & 0.297 & 0.276 & 77.26 & $5\times5, 5\times5, 5\times5$       & 0.362 & 0.408 & 77.80 \\
\bottomrule
\end{tabular}}
\end{adjustbox}
}
\vspace{-0.2cm}
\caption{Effect of multiple kernels in the depth-wise convolution of MKDC on BUSI dataset. The results for kernels beyond $7\times7$ are not reported as the performance does not scale proportionally with the computational cost of larger kernels. We use the MK-UNet network with \#channels= $[16,32,64,96,160]$ for these experiments and report the FLOPs for $256\times256$ inputs. We report the DICE scores (\%) averaging over five runs. Best results are highlighted in \textbf{bold}.}
\label{tab:multi_kernels}
\vspace{-0.2cm}
\end{table*}

\begin{table*}[t]
  \centering
    \begin{adjustbox}{width=0.65\textwidth}
\begin{tabular}{c|r|r|r|r|r|r|r|r}
\toprule
Blocks     & \#Params & \#FLOPs & BUSI & Clinic & Colon & ISIC18 & DSB18 & EM                    \\
\midrule
IRB          & 0.271M     & 0.230G     & 72.41  & 90.90 & 84.15 & 87.20 & 90.52 & 94.87  \\
MKIR (\textbf{Ours})         & 0.306M    & 0.300G     & \textbf{74.74} & \textbf{92.63} & \textbf{86.46} & \textbf{88.22} & \textbf{92.40} & \textbf{95.31}         \\
\bottomrule 
\end{tabular}
\end{adjustbox}
\vspace{-0.2cm}
\caption{Original Inverted Residual Block (IRB) \cite{sandler2018mobilenetv2} vs our Multi-Kernel Inverted Residual (MKIR) with \#channels = $[16,32,64,96,160]$. We use the kernel size of $[3]$ and $[1,3,5]$ for IRB and MKIR, respectively. We report the DICE scores (\%) averaging over five runs. Best results are shown in \textbf{bold}.}
\vspace{-0.2cm}
\label{tab:ablation_irb_vs_mkir}
\end{table*}

\begin{table*}[!h]
\begin{center}
    \begin{adjustbox}{width=0.7\textwidth}
\begin{tabular}{c|c|r|r|r|r|r|r|r|r}
\toprule
Encoder  & Decoder   & \#Params & \#FLOPs & BUSI & Clinic & Colon & ISIC18 & DSB18 & EM                    \\
\midrule
MKIRA & MKIRA         & 0.321M    & 0.346G    &  77.28 & 92.81 & 89.63 &  88.61 & 92.65 & 95.43     \\
(\textbf{Ours}) MKIR & MKIRA        & 0.316M    & 0.314G    & \textbf{78.04} & \textbf{93.48} & \textbf{90.01} & \textbf{88.64} & \textbf{92.71} & \textbf{95.52}       \\
\bottomrule 
\end{tabular}
\end{adjustbox}
\vspace{-0.2cm}
\caption{Effect of MKIRA in the encoder and decoder of MK-UNet with \#channels = $[16,32,64,96,160]$ and $[1,3,5]$ kernels. We report the DICE scores (\%) averaging over five runs. Best results are shown in bold.}
\label{tab:ablation_mkir_vs_mkira_encoder}
\end{center}
\vspace{-0.5cm}
\end{table*}

\subsection{Qualitative results}
\label{ssec:qualitative_results}

In Figure \ref{fig:qualitative}, we report the segmentation maps of breast tumors, skin lesions, polyps, and cell segmentation for representative test images. In breast tumor segmentation, UNet, UNet++, and UNeXt show greater false segmentation, while TransUNet and our MK-UNet produce near-perfect segmentation maps. Similarly, in skin lesion segmentation, UNet, ResUNet, UNet++, AttnUNet, DeepLabV3+, PraNet, SwinuNet, and UNeXt miss part of the lesion (in red rectangular box). However, UACANet, TransUNet, ACC-UNet, and our MK-UNet can segment that challenging region well. Our MK-UNet can also segment the polyp correctly, while all other methods incorrectly segment another region as a polyp. In general, our MK-UNet produces the best overlapping segmentation map in all four tasks. The reason behind this well-rounded performance by our MK-UNet with a very low computational budget is the use of multi-kernel depth-wise convolutions along with gated and local attention mechanisms.

%% file: sections/4.ablations.tex
\section{Ablation Study}
\label{sec:ablation_study}
This section describes three critical ablation studies. More ablation results are given in the Appendix.
\subsection{Effect of different components}
\label{ssec:components}
Table \ref{tab:ablation_components} presents the performance of various configurations within the MK-UNet network across six medical image segmentation datasets, highlighting the impact of integrating different components like MKIR, GAG, and MKIRA. The comparison spans models from UNeXt to the advanced MKIR + GAG + MKIRA variant, revealing a progressive improvement in the DICE scores with the addition of each component. Notably, the multi-kernel trick, implemented through MKIR (in the encoder) and MKIRA (in the decoder) blocks, is the most critical component for improving the segmentation accuracy, increasing the DICE score from 72.41\% to 76.61\% on the BUSI dataset. This indicates the significant contribution of the multi-kernel approach to feature extraction and refinement. However, when we integrate all proposed modules (MKIR + GAG + MKIRA), our model achieves the highest overall DICE score of 78.04\% on the same dataset, with minimal computational resources (0.316M \#Params and 0.314G \#FLOPs). This exhibits the efficacy of combining multi-kernel convolution with attention mechanisms within MK-UNet.

\subsection{Effect of multiple kernels}
\label{ssec:multiple_kernels}
Table \ref{tab:multi_kernels} evaluates the influence of different convolutional kernel combinations on the performance of MKDC within the MK-UNet network, specifically for the BUSI dataset. By experimenting with a variety of kernel sizes ranging from $1$ to $3,5,7$, it becomes evident that a mix of $1,3,5$ kernels stands out by achieving the best DICE score of 78.04\% with a moderate increase in computational resources (0.316M \#Params and 0.314G \#FLOPs). This finding highlights the effectiveness of a multi-scale kernel approach in capturing diverse feature representations, thus significantly improving segmentation accuracy without a substantial rise in computational demands. Drawing from these empirical findings, we opt for the kernel combination of $[1,3,5]$ across all our experiments.

\subsection{Effectiveness of our MKIR over IRB \cite{sandler2018mobilenetv2}}
\label{app:irb_mkir}

Table \ref{tab:ablation_irb_vs_mkir} reports the results of the original IRB of MobileNetv2 \cite{sandler2018mobilenetv2} and our proposed MKIR block. It can be concluded from the table that our MKIR significantly outperforms (up to 2.33\%) IRB in all the datasets with only an additional 0.035M \#Params and 0.07G \#FLOPs. The use of lightweight convolutions with multiple kernels contributes to these performance improvements with nominal additional computational resources.

\subsection{Effectiveness of MKIR vs. MKIRA in Encoder}
\label{assec:mkir_mkira}
Table \ref{tab:ablation_mkir_vs_mkira_encoder} demonstrates that employing MKIR in the encoder and MKIRA in the decoder yields superior performance across all datasets. Specifically, this configuration achieves the best average DICE scores of 78.04\% (BUSI), 93.48\% (Clinic), 90.01\% (Colon), 88.64\% (ISIC18), 92.71\% (DSB18), and 95.52\% (EM). The MKIR block in the encoder effectively extracts complex features by leveraging multiple kernels to capture a diverse range of spatial patterns and global contexts without the need for localized attention, which is more computationally intensive. Since the encoder primarily focuses on feature extraction, this design helps preserve critical details while maintaining lightweightness. In contrast, localized attention is crucial in the decoder to facilitate precise reconstruction. The MKIRA in the decoder attends to key spatial regions, enabling effective feature refinement. This complementary setup leads to an optimal balance between performance and computational cost, as evidenced by the superior results achieved with only 0.316M \#Params and 0.314G \#FLOPs.

%% file: sections/5.conclusion.tex
\section{Conclusion}
\label{sec:conclusion}
In this paper, we have presented MK-UNet, a significant advancement in the realm of medical image segmentation. MK-UNet addresses the long-standing challenge of balancing high segmentation accuracy with computational efficiency. By utilizing depth-wise convolution and multi-kernel processing, MK-UNet outperforms models like TransUNet and UNeXt with a fraction of their computational overhead (nearly 333$\times$ lower \#Params and 123$\times$ less complexity than TransUNet, and 4.7$\times$ fewer \#Params compared to UNeXt). This significant reduction in computational resources, coupled with improved performance, makes MK-UNet an optimal choice for deployment in environments where computational resources are limited, such as in point-of-care devices. 

The current evaluations of our proposed MK-UNet architecture are limited to binary medical segmentation tasks. In the future, we will extend experiments on multi-class and 3D segmentation tasks.

%% file: sections/6.supply.tex
\clearpage
\setcounter{page}{1}
\maketitlesupplementary

\subsection{Analysis of the number of channels}
\label{assec:number_of_channels}
We conduct an ablation study with the different number of channel dimensions in different stages of the network to show the scalability of our network.
Table \ref{tab:ablation_channels} reports the results of this set of experiments. The progression from MK-UNet-T to MK-UNet-L in Table \ref{tab:ablation_channels} demonstrates a clear positive correlation between model complexity and performance. Starting with MK-UNet-T's minimal resource use (0.027M \#Params, 0.062G \#FLOPs) yielding a 75.64\% DICE score on BUSI, the score increases to 78.04\% with MK-UNet's moderate complexity (0.316M \#Params, 0.314G \#FLOPs), and peaks at 79.02\% with MK-UNet-L's higher resource demand (3.76M \#Params, 3.19G \#FLOPs). This trend of increasing DICE score with model complexity is consistent across datasets.

\begin{table*}[!b]
\begin{center}
    \begin{adjustbox}{width=0.9\textwidth}
\begin{tabular}{c|r|r|r|r|r|r|r|r|r|r|r|r|r|r}
\toprule
Network     & C1    & C2 & C3 & C4 & C5  & \#Params & \#FLOPs & BUSI & Clinic & Colon & ISIC18 & DSB18 & EM                  \\
\midrule
MK-UNet-T           & 4  & 8 & 16  &  24 & 32     &  0.027M  & 0.062G & 75.64 & 91.26 &  85.03 & 88.19 & 92.38 & 94.69 \\
MK-UNet-S          & 8  & 16  & 32   & 48 & 80     &  0.093M & 0.125G & 77.26 & 92.31 & 88.78 & 88.57 & 92.45 & 95.22 \\
MK-UNet          & 16 & 32 & 64 & 96   & 160     & 0.316M & 0.314G &  78.04 & 93.48 &  90.01 & 88.74 & 92.71 & 95.52 \\
MK-UNet-M          & 32  & 64  & 128  &  192   & 320    & 1.15M  & 0.951G &  78.27 & 93.67 & 90.27 & 89.08 & 92.74 & 95.62 \\
MK-UNet-L          & 64  & 128  & 256 & 384  & 512    & 3.76M & 3.19G &  79.02 & 93.85 & 91.82 & 89.25 & 92.80 & 95.67 \\
\bottomrule 
\end{tabular}
\end{adjustbox}
\caption{Analysis of the number of channels on different datasets. \#FLOPs are reported for $256\times256$ inputs. We report the DICE scores (\%) averaging over five runs, thus having 1-4\% standard deviations.}
\label{tab:ablation_channels}
\end{center}
\end{table*}

\subsection{Effectiveness of our Grouped Attention Gate (GAG) over Attention Gate (AG) \cite{oktay2018attention}}
\label{assec:ag_gag}

Table \ref{tab:ablation_ag_vs_gag} reports the results of the original AG of Attention UNet \cite{oktay2018attention} and our proposed GAG block. It can be seen from the table that our GAG surpasses AG in all datasets with 0.01M fewer \#Params and 0.06G less \#FLOPs. The use of group convolutions with a relatively larger kernel ($3$) contributes to these performance improvements with less computational costs.

\begin{table*}[!b]
\centering    
    \begin{adjustbox}{width=0.7\textwidth}
\begin{tabular}{c|r|r|r|r|r|r|r|r}
\toprule
Blocks     & \#Params & \#FLOPs & BUSI & Clinic & Colon & ISIC18 & DSB18 & EM                    \\
\midrule
AG          & 0.326M     & 0.320G     & 77.61  & 93.02 & 89.78 & 88.38 & 92.48 & 95.31  \\
GAG (\textbf{Ours})         & 0.316M    & 0.314G     & \textbf{78.04} & \textbf{93.48} & \textbf{90.01} & \textbf{88.64} & \textbf{92.71} & \textbf{95.52}         \\
\bottomrule 
\end{tabular}
\end{adjustbox}
\caption{Original Attention Gate (AG) \cite{sandler2018mobilenetv2} vs our Grouped Attention Gate (GAG) with \#channels = $[16,32,64,96,160]$ in MK-UNet. We use the kernel size of $3$ for GAG. We report the DICE scores (\%) averaging over five runs. Best results are shown in bold.}
\vspace{-0.4cm}
\label{tab:ablation_ag_vs_gag}
\end{table*}


%% file: main.bbl
\begin{thebibliography}{44}
\providecommand{\natexlab}[1]{#1}
\providecommand{\url}[1]{\texttt{#1}}
\expandafter\ifx\csname urlstyle\endcsname\relax
  \providecommand{\doi}[1]{doi: #1}\else
  \providecommand{\doi}{doi: \begingroup \urlstyle{rm}\Url}\fi

\bibitem[Al-Dhabyani et~al.(2020)Al-Dhabyani, Gomaa, Khaled, and Fahmy]{al2020dataset}
Walid Al-Dhabyani, Mohammed Gomaa, Hussien Khaled, and Aly Fahmy.
\newblock Dataset of breast ultrasound images.
\newblock \emph{Data in brief}, 28:\penalty0 104863, 2020.

\bibitem[Bernal et~al.(2015)Bernal, S{\'a}nchez, Fern{\'a}ndez-Esparrach, Gil, Rodr{\'\i}guez, and Vilari{\~n}o]{bernal2015wm}
Jorge Bernal, F~Javier S{\'a}nchez, Gloria Fern{\'a}ndez-Esparrach, Debora Gil, Cristina Rodr{\'\i}guez, and Fernando Vilari{\~n}o.
\newblock Wm-dova maps for accurate polyp highlighting in colonoscopy: Validation vs. saliency maps from physicians.
\newblock \emph{Computerized medical imaging and graphics}, 43:\penalty0 99--111, 2015.

\bibitem[Caicedo et~al.(2019)Caicedo, Goodman, Karhohs, Cimini, Ackerman, Haghighi, Heng, Becker, Doan, McQuin, et~al.]{caicedo2019nucleus}
Juan~C Caicedo, Allen Goodman, Kyle~W Karhohs, Beth~A Cimini, Jeanelle Ackerman, Marzieh Haghighi, CherKeng Heng, Tim Becker, Minh Doan, Claire McQuin, et~al.
\newblock Nucleus segmentation across imaging experiments: the 2018 data science bowl.
\newblock \emph{Nature methods}, 16\penalty0 (12):\penalty0 1247--1253, 2019.

\bibitem[Cao et~al.(2022)Cao, Wang, Chen, Jiang, Zhang, Tian, and Wang]{cao2021swin}
Hu Cao, Yueyue Wang, Joy Chen, Dongsheng Jiang, Xiaopeng Zhang, Qi Tian, and Manning Wang.
\newblock Swin-unet: Unet-like pure transformer for medical image segmentation.
\newblock In \emph{Proceedings of the European Conference on Computer Vision Workshops}, pages 205--218. Springer, 2022.

\bibitem[Cardona et~al.(2010)Cardona, Saalfeld, Preibisch, Schmid, Cheng, Pulokas, Tomancak, and Hartenstein]{cardona2010integrated}
Albert Cardona, Stephan Saalfeld, Stephan Preibisch, Benjamin Schmid, Anchi Cheng, Jim Pulokas, Pavel Tomancak, and Volker Hartenstein.
\newblock An integrated micro-and macroarchitectural analysis of the drosophila brain by computer-assisted serial section electron microscopy.
\newblock \emph{PLoS biology}, 8\penalty0 (10):\penalty0 e1000502, 2010.

\bibitem[Chen et~al.(2024)Chen, Mei, Li, Lu, Yu, Wei, Luo, Xie, Adeli, Wang, et~al.]{chen2021transunet}
Jieneng Chen, Jieru Mei, Xianhang Li, Yongyi Lu, Qihang Yu, Qingyue Wei, Xiangde Luo, Yutong Xie, Ehsan Adeli, Yan Wang, et~al.
\newblock Transunet: Rethinking the u-net architecture design for medical image segmentation through the lens of transformers.
\newblock \emph{Medical Image Analysis}, page 103280, 2024.

\bibitem[Chen et~al.(2017{\natexlab{a}})Chen, Zhang, Xiao, Nie, Shao, Liu, and Chua]{chen2017sca}
Long Chen, Hanwang Zhang, Jun Xiao, Liqiang Nie, Jian Shao, Wei Liu, and Tat-Seng Chua.
\newblock Sca-cnn: Spatial and channel-wise attention in convolutional networks for image captioning.
\newblock In \emph{Proceedings of the IEEE/CVF Conference on Computer Vision and Pattern Recognition}, pages 5659--5667, 2017{\natexlab{a}}.

\bibitem[Chen et~al.(2017{\natexlab{b}})Chen, Papandreou, Kokkinos, Murphy, and Yuille]{chen2017deeplab}
Liang-Chieh Chen, George Papandreou, Iasonas Kokkinos, Kevin Murphy, and Alan~L Yuille.
\newblock Deeplab: Semantic image segmentation with deep convolutional nets, atrous convolution, and fully connected crfs.
\newblock \emph{IEEE Transactions on Pattern Analysis and Machine Intelligence}, 40\penalty0 (4):\penalty0 834--848, 2017{\natexlab{b}}.

\bibitem[Codella et~al.(2019)Codella, Rotemberg, Tschandl, Celebi, Dusza, Gutman, Helba, Kalloo, Liopyris, Marchetti, et~al.]{codella2019skin}
Noel Codella, Veronica Rotemberg, Philipp Tschandl, M~Emre Celebi, Stephen Dusza, David Gutman, Brian Helba, Aadi Kalloo, Konstantinos Liopyris, Michael Marchetti, et~al.
\newblock Skin lesion analysis toward melanoma detection 2018: A challenge hosted by the international skin imaging collaboration (isic).
\newblock \emph{arXiv preprint arXiv:1902.03368}, 2019.

\bibitem[Codella et~al.(2018)Codella, Gutman, Celebi, Helba, Marchetti, Dusza, Kalloo, Liopyris, Mishra, Kittler, et~al.]{codella2018skin}
Noel~CF Codella, David Gutman, M~Emre Celebi, Brian Helba, Michael~A Marchetti, Stephen~W Dusza, Aadi Kalloo, Konstantinos Liopyris, Nabin Mishra, Harald Kittler, et~al.
\newblock Skin lesion analysis toward melanoma detection: A challenge at the 2017 international symposium on biomedical imaging (isbi), hosted by the international skin imaging collaboration (isic).
\newblock In \emph{IEEE International Symposium on Biomedical Imaging}, pages 168--172. IEEE, 2018.

\bibitem[Dosovitskiy et~al.(2020)Dosovitskiy, Beyer, Kolesnikov, Weissenborn, Zhai, Unterthiner, Dehghani, Minderer, Heigold, Gelly, et~al.]{dosovitskiy2020image}
Alexey Dosovitskiy, Lucas Beyer, Alexander Kolesnikov, Dirk Weissenborn, Xiaohua Zhai, Thomas Unterthiner, Mostafa Dehghani, Matthias Minderer, Georg Heigold, Sylvain Gelly, et~al.
\newblock An image is worth 16x16 words: Transformers for image recognition at scale.
\newblock \emph{arXiv preprint arXiv:2010.11929}, 2020.

\bibitem[Fan et~al.(2020)Fan, Ji, Zhou, Chen, Fu, Shen, and Shao]{fan2020pranet}
Deng-Ping Fan, Ge-Peng Ji, Tao Zhou, Geng Chen, Huazhu Fu, Jianbing Shen, and Ling Shao.
\newblock Pranet: Parallel reverse attention network for polyp segmentation.
\newblock In \emph{International Conference on Medical Image Computing and Computer-Assisted Intervention}, pages 263--273. Springer, 2020.

\bibitem[He et~al.(2016)He, Zhang, Ren, and Sun]{he2016deep}
Kaiming He, Xiangyu Zhang, Shaoqing Ren, and Jian Sun.
\newblock Deep residual learning for image recognition.
\newblock In \emph{Proceedings of the IEEE/CVF Conference on Computer Vision and Pattern Recognition}, pages 770--778, 2016.

\bibitem[Howard et~al.(2017)Howard, Zhu, Chen, Kalenichenko, Wang, Weyand, Andreetto, and Adam]{howard2017mobilenets}
Andrew~G Howard, Menglong Zhu, Bo Chen, Dmitry Kalenichenko, Weijun Wang, Tobias Weyand, Marco Andreetto, and Hartwig Adam.
\newblock Mobilenets: Efficient convolutional neural networks for mobile vision applications.
\newblock \emph{arXiv preprint arXiv:1704.04861}, 2017.

\bibitem[Hu et~al.(2018)Hu, Shen, and Sun]{hu2018squeeze}
Jie Hu, Li Shen, and Gang Sun.
\newblock Squeeze-and-excitation networks.
\newblock In \emph{Proceedings of the IEEE/CVF Conference on Computer Vision and Pattern Recognition}, pages 7132--7141, 2018.

\bibitem[Ibtehaz and Kihara(2023)]{ibtehaz2023acc}
Nabil Ibtehaz and Daisuke Kihara.
\newblock Acc-unet: A completely convolutional unet model for the 2020s.
\newblock In \emph{International Conference on Medical Image Computing and Computer-Assisted Intervention}, pages 692--702. Springer, 2023.

\bibitem[Ioffe and Szegedy(2015)]{ioffe2015batch}
Sergey Ioffe and Christian Szegedy.
\newblock Batch normalization: Accelerating deep network training by reducing internal covariate shift.
\newblock In \emph{International Conference on Machine Learning}, pages 448--456. pmlr, 2015.

\bibitem[Kim et~al.(2021)Kim, Lee, and Kim]{kim2021uacanet}
Taehun Kim, Hyemin Lee, and Daijin Kim.
\newblock Uacanet: Uncertainty augmented context attention for polyp segmentation.
\newblock In \emph{ACM International Conference on Multimedia}, pages 2167--2175, 2021.

\bibitem[Krizhevsky and Hinton(2010)]{krizhevsky2010convolutional}
Alex Krizhevsky and Geoff Hinton.
\newblock Convolutional deep belief networks on cifar-10.
\newblock \emph{Unpublished manuscript}, 40\penalty0 (7):\penalty0 1--9, 2010.

\bibitem[Lin et~al.(2023)Lin, Wu, Chen, Huang, and Jin]{lin2023scale}
Weifeng Lin, Ziheng Wu, Jiayu Chen, Jun Huang, and Lianwen Jin.
\newblock Scale-aware modulation meet transformer.
\newblock In \emph{Proceedings of the IEEE/CVF International Conference on Computer Vision}, pages 6015--6026, 2023.

\bibitem[Liu et~al.(2024)Liu, Zhu, Liu, Yu, Chen, and Gao]{liu2024rolling}
Yutong Liu, Haijiang Zhu, Mengting Liu, Huaiyuan Yu, Zihan Chen, and Jie Gao.
\newblock Rolling-unet: Revitalizing mlp’s ability to efficiently extract long-distance dependencies for medical image segmentation.
\newblock In \emph{Proceedings of the AAAI Conference on Artificial Intelligence}, pages 3819--3827, 2024.

\bibitem[Liu et~al.(2021)Liu, Lin, Cao, Hu, Wei, Zhang, Lin, and Guo]{liu2021swin}
Ze Liu, Yutong Lin, Yue Cao, Han Hu, Yixuan Wei, Zheng Zhang, Stephen Lin, and Baining Guo.
\newblock Swin transformer: Hierarchical vision transformer using shifted windows.
\newblock In \emph{Proceedings of the IEEE/CVF International Conference on Computer Vision}, pages 10012--10022, 2021.

\bibitem[Long et~al.(2015)Long, Shelhamer, and Darrell]{long2015fully}
Jonathan Long, Evan Shelhamer, and Trevor Darrell.
\newblock Fully convolutional networks for semantic segmentation.
\newblock In \emph{Proceedings of the IEEE/CVF Conference on Computer Vision and Pattern Recognition}, pages 3431--3440, 2015.

\bibitem[Loshchilov and Hutter(2017)]{loshchilov2017decoupled}
Ilya Loshchilov and Frank Hutter.
\newblock Decoupled weight decay regularization.
\newblock \emph{arXiv preprint arXiv:1711.05101}, 2017.

\bibitem[Nair and Hinton(2010)]{nair2010rectified}
Vinod Nair and Geoffrey~E Hinton.
\newblock Rectified linear units improve restricted boltzmann machines.
\newblock In \emph{International Conference on Machine Learning}, pages 807--814, 2010.

\bibitem[Oktay et~al.(2018)Oktay, Schlemper, Folgoc, Lee, Heinrich, Misawa, Mori, McDonagh, Hammerla, Kainz, et~al.]{oktay2018attention}
Ozan Oktay, Jo Schlemper, Loic~Le Folgoc, Matthew Lee, Mattias Heinrich, Kazunari Misawa, Kensaku Mori, Steven McDonagh, Nils~Y Hammerla, Bernhard Kainz, et~al.
\newblock Attention u-net: Learning where to look for the pancreas.
\newblock In \emph{Medical Imaging with Deep Learning}, 2018.

\bibitem[Rahman and Marculescu(2023{\natexlab{a}})]{Rahman_2023_WACV}
Md~Mostafijur Rahman and Radu Marculescu.
\newblock Medical image segmentation via cascaded attention decoding.
\newblock In \emph{Proceedings of the IEEE/CVF Winter Conference on Applications of Computer Vision}, pages 6222--6231, 2023{\natexlab{a}}.

\bibitem[Rahman and Marculescu(2023{\natexlab{b}})]{rahman2023multi}
Md~Mostafijur Rahman and Radu Marculescu.
\newblock Multi-scale hierarchical vision transformer with cascaded attention decoding for medical image segmentation.
\newblock In \emph{Medical Imaging with Deep Learning}, 2023{\natexlab{b}}.

\bibitem[Rahman and Marculescu(2024)]{rahman2024g}
Md~Mostafijur Rahman and Radu Marculescu.
\newblock G-cascade: Efficient cascaded graph convolutional decoding for 2d medical image segmentation.
\newblock In \emph{Proceedings of the IEEE/CVF Winter Conference on Applications of Computer Vision}, pages 7728--7737, 2024.

\bibitem[Rahman et~al.(2024)Rahman, Munir, and Marculescu]{rahman2024emcad}
Md~Mostafijur Rahman, Mustafa Munir, and Radu Marculescu.
\newblock Emcad: Efficient multi-scale convolutional attention decoding for medical image segmentation.
\newblock In \emph{Proceedings of the IEEE/CVF Conference on Computer Vision and Pattern Recognition}, pages 11769--11779, 2024.

\bibitem[Ronneberger et~al.(2015)Ronneberger, Fischer, and Brox]{ronneberger2015u}
Olaf Ronneberger, Philipp Fischer, and Thomas Brox.
\newblock U-net: Convolutional networks for biomedical image segmentation.
\newblock In \emph{International Conference on Medical Image Computing and Computer-Assisted Intervention}, pages 234--241. Springer, 2015.

\bibitem[Ruan et~al.(2022)Ruan, Xiang, Xie, Liu, and Fu]{ruan2022malunet}
Jiacheng Ruan, Suncheng Xiang, Mingye Xie, Ting Liu, and Yuzhuo Fu.
\newblock Malunet: A multi-attention and light-weight unet for skin lesion segmentation.
\newblock In \emph{2022 IEEE International Conference on Bioinformatics and Biomedicine (BIBM)}, pages 1150--1156. IEEE, 2022.

\bibitem[Ruan et~al.(2023)Ruan, Xie, Gao, Liu, and Fu]{ruan2023ege}
Jiacheng Ruan, Mingye Xie, Jingsheng Gao, Ting Liu, and Yuzhuo Fu.
\newblock Ege-unet: an efficient group enhanced unet for skin lesion segmentation.
\newblock In \emph{International Conference on Medical Image Computing and Computer-Assisted Intervention}, pages 481--490. Springer, 2023.

\bibitem[Sandler et~al.(2018)Sandler, Howard, Zhu, Zhmoginov, and Chen]{sandler2018mobilenetv2}
Mark Sandler, Andrew Howard, Menglong Zhu, Andrey Zhmoginov, and Liang-Chieh Chen.
\newblock Mobilenetv2: Inverted residuals and linear bottlenecks.
\newblock In \emph{Proceedings of the IEEE/CVF Conference on Computer Vision and Pattern Recognition}, pages 4510--4520, 2018.

\bibitem[Seo et~al.(2022)Seo, So, Yun, Lee, and Barg]{seo2022spatial}
Hyunseok Seo, Seohee So, Sojin Yun, Seokjun Lee, and Jiseong Barg.
\newblock Spatial feature conservation networks (sfcns) for dilated convolutions to improve breast cancer segmentation from dce-mri.
\newblock In \emph{International Workshop on Applications of Medical AI}, pages 118--127. Springer, 2022.

\bibitem[Tan and Le(2019)]{tan2019efficientnet}
Mingxing Tan and Quoc Le.
\newblock Efficientnet: Rethinking model scaling for convolutional neural networks.
\newblock In \emph{International Conference on Machine Learning}, pages 6105--6114. PMLR, 2019.

\bibitem[Tang et~al.(2024)Tang, Ding, Quan, Wang, Ning, and Zhou]{tang2023cmunext}
Fenghe Tang, Jianrui Ding, Quan Quan, Lingtao Wang, Chunping Ning, and S~Kevin Zhou.
\newblock Cmunext: An efficient medical image segmentation network based on large kernel and skip fusion.
\newblock In \emph{IEEE International Symposium on Biomedical Imaging}, pages 1--5. IEEE, 2024.

\bibitem[Valanarasu and Patel(2022)]{valanarasu2022unext}
Jeya Maria~Jose Valanarasu and Vishal~M Patel.
\newblock Unext: Mlp-based rapid medical image segmentation network.
\newblock In \emph{International Conference on Medical Image Computing and Computer-Assisted Intervention}, pages 23--33. Springer, 2022.

\bibitem[Valanarasu et~al.(2021)Valanarasu, Oza, Hacihaliloglu, and Patel]{valanarasu2021medical}
Jeya Maria~Jose Valanarasu, Poojan Oza, Ilker Hacihaliloglu, and Vishal~M Patel.
\newblock Medical transformer: Gated axial-attention for medical image segmentation.
\newblock In \emph{International Conference on Medical Image Computing and Computer-Assisted Intervention}, pages 36--46. Springer, 2021.

\bibitem[V{\'a}zquez et~al.(2017)V{\'a}zquez, Bernal, S{\'a}nchez, Fern{\'a}ndez-Esparrach, L{\'o}pez, Romero, Drozdzal, and Courville]{vazquez2017benchmark}
David V{\'a}zquez, Jorge Bernal, F~Javier S{\'a}nchez, Gloria Fern{\'a}ndez-Esparrach, Antonio~M L{\'o}pez, Adriana Romero, Michal Drozdzal, and Aaron Courville.
\newblock A benchmark for endoluminal scene segmentation of colonoscopy images.
\newblock \emph{Journal of Healthcare Engineering}, 2017, 2017.

\bibitem[Woo et~al.(2018)Woo, Park, Lee, and Kweon]{woo2018cbam}
Sanghyun Woo, Jongchan Park, Joon-Young Lee, and In~So Kweon.
\newblock Cbam: Convolutional block attention module.
\newblock In \emph{International Conference on Machine Learning}, pages 3--19, 2018.

\bibitem[Wu et~al.(2024)Wu, Liu, Liang, and Chang]{wu2024ultralight}
Renkai Wu, Yinghao Liu, Pengchen Liang, and Qing Chang.
\newblock Ultralight vm-unet: Parallel vision mamba significantly reduces parameters for skin lesion segmentation.
\newblock \emph{arXiv preprint arXiv:2403.20035}, 2024.

\bibitem[Zhang et~al.(2018)Zhang, Liu, and Wang]{zhang2018road}
Zhengxin Zhang, Qingjie Liu, and Yunhong Wang.
\newblock Road extraction by deep residual u-net.
\newblock \emph{IEEE Geoscience and Remote Sensing Letters}, 15\penalty0 (5):\penalty0 749--753, 2018.

\bibitem[Zhou et~al.(2018)Zhou, Rahman~Siddiquee, Tajbakhsh, and Liang]{zhou2018unet++}
Zongwei Zhou, Md~Mahfuzur Rahman~Siddiquee, Nima Tajbakhsh, and Jianming Liang.
\newblock Unet++: A nested u-net architecture for medical image segmentation.
\newblock In \emph{International Workshop on Deep Learning in Medical Image Analysis and Multimodal Learning for Clinical Decision Support}, pages 3--11. Springer, 2018.

\end{thebibliography}
